\documentclass{article}

\usepackage{microtype}
\usepackage{graphicx}
\usepackage{subfigure}
\usepackage{booktabs} 

\usepackage{hyperref}
\usepackage{url}
\usepackage{graphicx}
\usepackage{pdfpages}
\usepackage{amsfonts}
\usepackage{dsfont}
\usepackage{xcolor}
\usepackage{bm}

\usepackage{hyperref}
\usepackage{amsmath}
\usepackage{amssymb}

\usepackage{enumitem}

\usepackage{hyperref}

\usepackage[accepted]{icml2018}

\icmltitlerunning{Testing with Concept Activation Vectors (TCAV)}

\newcommand{\TCAVscore}{{\mbox{TCAV}{\mbox{\small\hspace*{-0.1em}\scshape q}}}}
\begin{document}

\twocolumn[
\icmltitle{Interpretability Beyond Feature Attribution: \\ Quantitative Testing with Concept Activation Vectors (TCAV)}




\icmlsetsymbol{equal}{*}

\begin{icmlauthorlist}
\icmlauthor{Been Kim}{}
\icmlauthor{Martin Wattenberg}{}
\icmlauthor{Justin Gilmer}{}
\icmlauthor{Carrie Cai}{} 
\icmlauthor{James Wexler}{} \\
\icmlauthor{Fernanda Viegas }{}
\icmlauthor{Rory Sayres}{}
\end{icmlauthorlist}


\icmlcorrespondingauthor{Been Kim}{beenkim@google.com}

\icmlkeywords{Machine Learning, ICML}

\vskip 0.3in
]




\begin{abstract}
The interpretation of deep learning models is a challenge due to their size, complexity, and often opaque internal state. In addition, many systems, such as image classifiers, operate on low-level features rather than high-level concepts. To address these challenges, we introduce Concept Activation Vectors (CAVs), which provide an interpretation of a neural net's internal state in terms of human-friendly concepts. The key idea is to view the high-dimensional internal state of a neural net as an aid, not an obstacle. We show how to use CAVs as part of a technique, Testing with CAVs (TCAV), that uses directional derivatives to quantify the degree to which a user-defined concept is important to a classification result--for example, how sensitive a prediction of “zebra” is to the presence of stripes. Using the domain of image classification as a testing ground, we describe how CAVs may be used to explore hypotheses and generate insights for a standard image classification network as well as a medical application.
\end{abstract}

\section{Introduction}

Understanding the behavior of modern machine learning (ML) models, such as neural networks, remains a significant challenge. Given the breadth and importance of ML applications, however, it is important to address this challenge.  
In addition to ensuring accurate predictions,
and giving scientists and engineers better means of 
designing, developing, and debugging models,
interpretability is also important to ensure that ML models reflect our values.

One natural approach to interpretability is to describe an ML model's predictions in terms of the input features it considers. 
For instance, 
in logistic regression classifiers,
coefficient weights are often interpreted as the importance of each feature. 
Similarly, saliency maps give importance weights to pixels based on first-order derivatives~\citep{smilkov2017smoothgrad, selvaraju2016grad, sundararajan2017axiomatic, erhan2009visualizing, dabkowski2017real}. 

A key difficulty, however, is that most ML models operate on features, such as pixel values, that do not correspond to  high-level concepts that humans easily understand. Furthermore, a model's internal values (e.g., neural activations) can seem incomprehensible. We can express this difficulty mathematically, viewing the state of an ML model as a vector space $E_m$ spanned by basis vectors $e_m$ which correspond to data such as input features and neural activations.
Humans work in a different vector space $E_h$ spanned by implicit vectors $e_h$ corresponding to an unknown set of human-interpretable concepts.

From this standpoint, an ``interpretation'' of an ML model can be seen as function $g: E_m \rightarrow E_h$. When $g$ is linear, we call it a
\textbf{linear interpretability}.
In general, an interpretability function $g$ need not be perfect~\citep{DoshiKim2017Interpretability};
it may fail to explain some aspects of its input domain $E_m$
and it will unavoidably not cover all possible human concepts in $E_h$.

In this work, the high-level concepts of $E_h$ are defined using
sets of example input data for the ML model under inspection.
%
%
For instance, to define concept `curly', a set of hairstyles and texture images can be used. 
Note the concepts of $E_h$ are not constrained to input features or training data; they can be defined using new, user-provided data.
Examples are shown to be effective means of interfacing with ML models for both non-expert and expert users~\citep{koh2017understanding, kim2014bayesian, kim2015mind, klein1989decision}.

\begin{figure*}
\centering 
\includegraphics[width=.9 \linewidth]{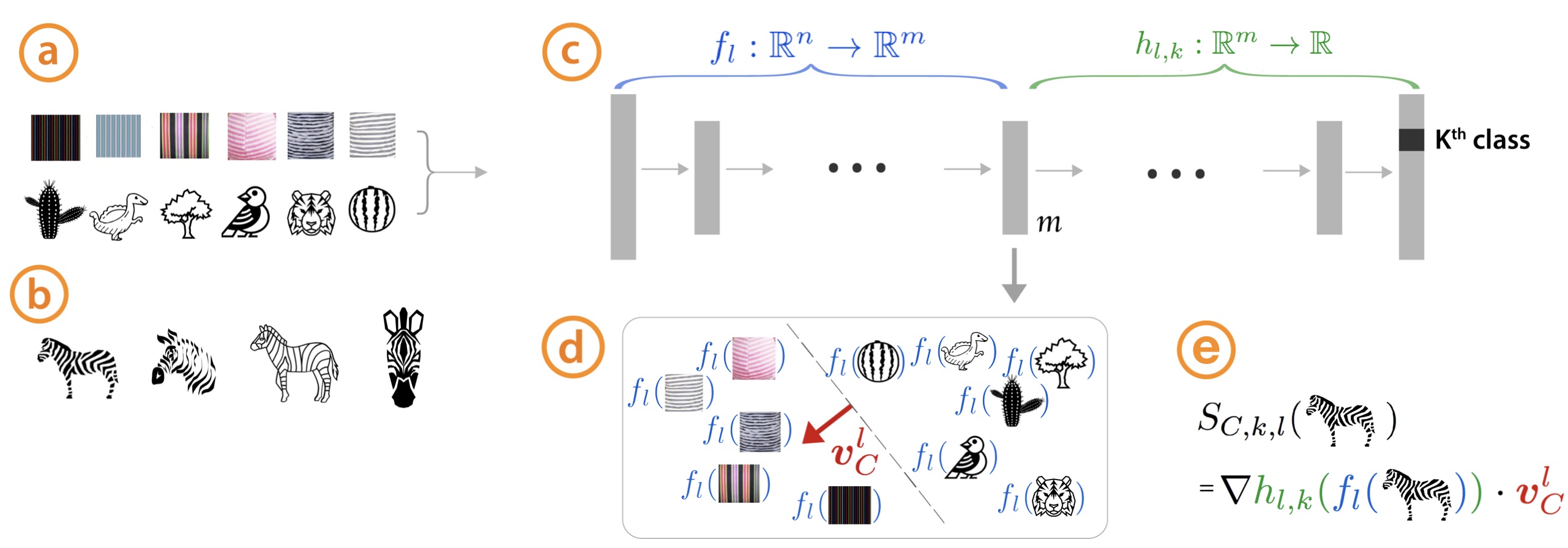}\vspace*{-1ex}
\caption{\textbf{Testing with Concept Activation Vectors:} Given
a user-defined set of examples for a concept (e.g., `striped'), and random examples \textcircled{a}, labeled training-data examples for the studied class (zebras) \textcircled{b}, and a trained network \textcircled{c}, TCAV can quantify the model's sensitivity to the concept for that class. CAVs are learned by training a linear classifier to distinguish between the activations produced by a concept's examples and examples in any layer \textcircled{d}. The CAV is the vector orthogonal to the classification boundary ($v_C^l$, red arrow). 
For the class of interest (zebras), TCAV uses the directional derivative $S_{C, k, l}(\bm{x})$ to quantify conceptual sensitivity \textcircled{e}. 
\label{fig:overview}}
\end{figure*}

This work introduces the notion of a \emph{Concept Activation Vector} (CAV) as a way of translating between $E_m$ and $E_h$.
A CAV for a concept is simply a vector in the direction of
the values (e.g., activations) of that concept's set of examples.
In this paper, we derive CAVs by training a linear classifier between a concept's examples and random counterexamples and then taking the vector orthogonal to the decision boundary. This simple approach is supported by recent work using local linearity~\cite{alain2016understanding, raghu2017svcca, netdissect2017, szegedy2013intriguing}.

The main result of this paper is a new linear interpretability method, quantitative \emph{Testing with CAV} (TCAV) (outlined in Figure~\ref{fig:overview}).
TCAV uses \emph{directional derivatives} to quantify the model prediction's sensitivity to an underlying high-level concept, learned by a CAV.
For instance, given an ML image model recognizing zebras, and a new, user-defined set of examples defining `striped', TCAV can quantify the influence of striped concept to the `zebra' prediction as a single number. 
In addition, we conduct statistical tests where
CAVs are randomly re-learned and rejected unless they show a significant and stable correlation with a model output class or state value. (This is detailed in Section~\ref{subsec:learnCAV}). 

Our work on TCAV was pursued with the following goals.
\begin{description}[topsep=-1ex,itemsep=-.5ex,partopsep=.5ex,parsep=1.0ex]
  \item \textbf{Accessibility:} Requires little to no ML expertise of user.
  \item \textbf{Customization:} Adapts to any concept (e.g., gender) and is not limited to concepts considered during training.
  \item \textbf{Plug-in readiness:} Works without any retraining or modification of the ML model.
  \item \textbf{Global quantification:} Can interpret entire classes or sets of examples with 
  a single quantitative measure, and not just explain individual data inputs.
\end{description}

We perform experiments using TCAV to gain insights and reveal dataset biases in widely-used neural network models and with a medical application (diabetic retinopathy), confirming our findings with a domain expert.
We conduct human subject experiments to quantitatively evaluate feature-based explanations and to contrast with TCAV.

\section{Related work\label{sec:related}}

In this section, we provide an overview of existing interpretability methods, methods specific to neural networks, and methods that leverage the local linearity of neural networks. 

\subsection{Interpretability methods}
To achieve interpretability, we have two options: (1) restrict ourselves to inherently interpretable models or (2) post-process our models in way that yields insights.
While option 1 offers simplicity as the explanation is embedded in the model~\citep{kim2014bayesian, doshi2015graph, lasso, sparsepca, ustun2013supersparse, caruana2015intelligible}, this option might be costly for users who already have a working high performance model.
With increasing demands for more explainable ML~\cite{goodman2016european}, there is an growing need for methods that can be applied without retraining or modifying the network. 

One of many challenges of option 2 
is to ensure that the explanation
correctly reflects the model's complex internals. 
One way to address this is to use the generated explanation as an input, and check the network's output for validation. This is typically used in perturbation-based/sensitivity analysis-based interpretability methods 
to either use data points~\citep{koh2017understanding} or features \cite{ribeiro2016should, LundbergL17} as a form of
perturbation, and check how the network's response changes. They maintain the consistency either locally
(i.e., explanation is true for a data point and its neighbors) or globally
(i.e., explanation is true for most data points in a class) by construction.
TCAV is a type of global perturbation method, as it perturbs data points towards a human-relatable concept to generate explanations.

However, even a perturbation-based method can be inconsistent 
if the explanation is only true for a particular data point and its neighbors~\citep{ribeiro2016should} (i.e., local explanation), and not for all inputs in the class. For example, they may generate
contradicting explanations for two data points in the same class, resulting in decreased user trust. TCAV produces explanations that are not only true for a single data point, but true for each class (i.e., global explanation).

\subsection{Interpretability methods in neural networks}
The goal of TCAV is to interpret high dimensional $E_m$ such as that of
neural network models. 
Saliency methods are one of the most popular local explanation methods for image classification~\citep{erhan2009visualizing, smilkov2017smoothgrad, selvaraju2016grad, sundararajan2017axiomatic, 
 dabkowski2017real}. These techniques typically produce a map showing how important each pixel of a particular picture
is for its classification, as shown in Figure~\ref{fig:sanity_sal}.
While a saliency map often identifies relevant regions and
provides a type of quantification (i.e., importance for each pixel), there are a couple of limitations:
1) since a saliency map is given conditioned on only \textit{one picture} 
(i.e., local explanation), humans have to manually assess each picture
in order to draw a class-wide conclusion, and
2) users have no control over what concepts of interest these maps pick up on (lack of customization).
For example, consider two saliency maps of two different cat pictures, with one picture's cat ears having more brightness. Can we assess how important the ears were in the prediction of ``cats''?

Furthermore, some recent work has demonstrated that saliency maps produced by randomized networks are similar to that of the trained network
~\cite{adebayo2018interpretation}, 
while simple meaningless data processing steps, such as mean shift,
may cause saliency methods to result in significant changes~\cite{saras}. Saliency maps may also be vulnerable
to adversarial attacks~\cite{ghorbani2017interpretation}.

\subsection{Linearity in neural network and latent dimensions}
There has been much research demonstrating that linear combinations of neurons 
may encode meaningful, insightful 
information~\cite{alain2016understanding, raghu2017svcca, netdissect2017, szegedy2013intriguing, Jesse17}.
Both \cite{netdissect2017} and \cite{alain2016understanding} show that meaningful directions
can be efficiently learned via simple linear classifiers. 
Mapping latent dimensions to human concepts has also been studied in the context of words~\cite{mikolov2013distributed}, 
and in the context of GANs to generate attribute-specific pictures~\cite{CycleGAN2017}.
A similar idea to using such concept vectors in latent dimension in the context of generative model has
also been explored~\cite{Jesse17}.

Our work extends this idea 
and computes directional derivatives along these learned directions in order to gather the importance of each direction for a model's prediction. Using TCAV's framework, we can
conduct hypothesis testing on any concept on the fly (customization) that make sense to the user (accessibility)
for a trained network (plug-in readiness) and produce a global explanation for each class.

\section{Methods}
This section explains our ideas and methods:
(a) how to use directional derivatives to quantify
the sensitivity of ML model predictions 
for different user-defined concepts,
and (b) how to compute
a final
quantitative explanation (\TCAVscore{} measure) of the relative importance of each concept to each model prediction class,
without any model retraining or modification.

Without loss of generality,
we consider neural network models
with inputs $\bm{x} \in \mathbb{R}^n$
and a feedforward layer $l$ with $m$ neurons, such that
input inference and its layer $l$ activations can be
seen as a function $f_l: \mathbb{R}^n \to \mathbb{R}^m$.

\subsection{User-defined Concepts as Sets of Examples}

The first step in our method is to define a concept of interest. We do this simply by choosing a set of examples that represent this concept or find an independent data set with the concept labeled. The key benefit of this strategy is that it 
does not restrict model interpretations
to explanations using
only pre-existing features, labels, or training data of the model under inspection.

Instead, there is great flexibility 
for even non-expert ML model analysts to define concepts
using examples and explore and refine concepts as they test hypotheses during analysis.
Section~\ref{sec:results} describes results from experiments with small number of images (30) collected using a search engine.
For the case of fairness analysis (e.g., gender, protected groups), 
curated examples are readily available~\cite{huang2007labeled}.

\subsection{Concept Activation Vectors (CAVs)\label{subsec:learnCAV}}

Following the approach of linear interpretability, given a set of examples representing a concept of human interest, we seek a vector in the space of activations of layer $l$ that represents this concept. To find such a vector, we consider the activations in layer $l$ produced by input examples that in the concept set versus random examples. We then define a ``concept activation vector'' (or CAV) as the normal to a hyperplane separating examples \emph{without} a concept and examples \emph{with} a concept
in the model's activations (see red arrow 
in Figure~\ref{fig:overview}).

This approach lends itself to a natural implementation. When an analyst is interested in a concept $C$ (say, striped textures) they may gather 
a positive set of example inputs $P_C$ (e.g., photos of striped objects) and 
and a negative set $N$ (e.g., a set of random photos).
Then, a binary linear classifier can be trained to distinguish between the
layer activations of the two sets:
$\{f_l(\bm{x}) : \bm{x} \in P_C\}$ and
$\{f_l(\bm{x}) : \bm{x} \in N\}$.%
\footnote{For convnets, a layer must be flattened so width $w$, height $h$, and $c$ channels becomes a vector of $m = w \times h \times c$ activations.}
This classifier $\bm{v}^l_C \in \mathbb{R}^m$ is a linear CAV for the concept $C$.

\subsection{Directional Derivatives and Conceptual Sensitivity}
Interpretability methods like saliency maps use the gradients of logit values with respect to individual input features, like pixels, and compute\vspace*{-2.0ex}
\[ \vspace*{-1.0ex}
\frac{\partial h_k(\bm{x}) }{\partial \bm{x}_{a,b}}
\]
where  $h_k(\bm{x})$ is the logit for a data point $\bm{x}$ for class $k$ and $\bm{x}_{a,b}$ is a pixel at position $(a,b)$ in $\bm{x}$. 
Thus, saliency uses the derivative
to gauge the sensitivity of the output class $k$
to changes in the magnitude of pixel $(a,b)$.

By using CAVs and directional derivatives,
we instead gauge the sensitivity of ML predictions
to changes in inputs towards the direction of a concept,
at neural activation layer $l$.
If $\bm{v}_C^l \in \mathbb{R}^m$ is a unit CAV vector for a concept $C$ in layer $l$, and $f_l(\bm{x})$ the activations for input $\bm{x}$ at layer $l$, the ``conceptual sensitivity'' of class $k$ to concept $C$ can be computed as the directional derivative  $S_{C,k,l}(\bm{x})$:
\vspace*{-1.3ex}
\begin{eqnarray}
S_{C,k,l}(\bm{x}) & = &
\lim\limits_{\epsilon \rightarrow 0} \frac{h_{l, k}(f_l(\bm{x}) + \epsilon \bm{v}_C^l) - h_{l,k}(f_l(\bm{x}))}{\epsilon} 
\nonumber \\
& = & \nabla h_{l,k}(f_l(\bm{x})) \cdot \bm{v}_C^l ,
\vspace*{-2ex}
\end{eqnarray}
where $h_{l, k}: \mathbb{R}^m \to \mathbb{R}$. 
This
$S_{C,k,l}(\bm{x})$  
can quantitatively measure the
sensitivity of model predictions with respect to concepts at any model layer.
It
is not a per-feature metric (e.g., unlike per-pixel saliency maps) 
but a per-concept scalar quantity computed on a whole input or sets of inputs.

\subsection{Testing with CAVs (TCAV)}\label{sec:testing}
Testing with CAVs, or TCAV, uses
directional derivatives
to compute ML models' conceptual sensitivity 
across entire classes of inputs.
Let $k$ be a class label for a given supervised learning task and let $X_k$ denote all inputs with that given label. We define the TCAV score to be 
\vspace*{-0.5ex}
\begin{equation}
 \TCAVscore_{C, k, l} = \frac{\left\lvert \left\{ 
                    \bm{x}\in X_k : 
                    S_{C,k,l}(\bm{x}) > 0 
                    \right\} \right\rvert}%
                   {\left\lvert X_k \right\rvert}
\vspace*{-0.5ex}
\end{equation}
i.e. the fraction of $k$-class inputs whose $l$-layer activation vector was positively influenced by concept $C$, $\TCAVscore_{C,k,l} \in [0, 1]$. Note that $\TCAVscore_{C,k,l}$ only depends on the sign of $S_{C,k,l}$, one could also use a different metric that considers the magnitude of the conceptual sensitivities. 
The \TCAVscore{} metric
allows 
conceptual sensitivities
to be 
easily interpreted, globally for all inputs in a label. 
%

\subsection{Statistical significance testing}\label{sec:significance}

One pitfall with the TCAV technique is the potential for learning a meaningless CAV. After all, using a randomly chosen set of images will still produce a CAV. A test based on such a random concept is unlikely to be meaningful.

To guard against spurious results from testing a class against a particular CAV, we propose the following simple statistical significance test. Instead of training a CAV once, against a single batch of random examples $N$, we perform multiple training runs, typically 500. A meaningful concept should lead to TCAV scores that behave consistently across training runs.

Concretely we perform a two-sided $t$-test of the TCAV scores based on these multiple samples. If we can reject the null hypothesis of a TCAV score of 0.5, we can consider the resulting concept as related to the class prediction in a significant way. Note that we also perform a Bonferroni correction for our hypotheses (at $p < \alpha/m$ with $m=2$) to control the false discovery rate further. All results shown in this paper are CAVs that passed this testing.
%


\subsection{TCAV extensions: Relative TCAV}
\label{subsec:relativeCAV}
In practice, semantically related concepts (e.g., brown hair vs. black hair) often yield CAVs that are far from orthogonal. This natural, expected property may be beneficially used to make fine-grained distinctions 
since relative comparisons between related concepts
are a good interpretative tool~\citep{kim2015mind, doshi2015graph, lasso, salvatore2014machine}.

\textit{Relative CAVs} allow making such fine-grained comparisons.
Here the analyst selects two sets of inputs that represent two different concepts, $C$ and $D$. Training a classifier on $f_l(P_C)$ and $f_l(P_D)$ yields a vector $\bm{v}^l_{C,D} \in \mathbb{R}^m$. The vector $\bm{v}^l_{C,D}$ intuitively defines a $1$-$d$ subspace in layer $l$ where the projection of an embedding $f_l(\bm{x})$ along this subspace measures whether $\bm{x}$ is more relevant to concept $C$ or $D$.

Relative CAVs may, for example, apply to image recognition,
where we can hypothesize that
concepts for `dotted', `striped', and `meshed' textures 
are likely to exist as internal representations, 
and be correlated or overlapping.
Given three positive example sets $P_\textup{dot}$, $P_\textup{stripe}$, and $P_\textup{mesh}$,
a relative CAV can be derived by constructing, for each,
a negative input set by complement
(e.g., $\left\{ P_\textup{dot} \cup P_\textup{mesh} \right\}$ for the stripes).
The \TCAVscore{} measures
enabled by the resulting relative CAV
are used in many of the experiments 
in the following Section~\ref{sec:results},
e.g., to gauge
the relative importance of stripes to zebras and
that of diagnostic concepts for diabetic retinopathy.

\section{Results\label{sec:results}}

We first show evidence that CAVs align with intended concepts of interest, by sorting images based on how similar they are to various concepts (Section~\ref{subsec:sorter})
and by using an activation maximization technique, \textit{empirical deep dream}, on the CAVs (Section~\ref{subsec:deepdream}).
We then summarize gained insights and revealed biases of two widely used networks using TCAV (Section~\ref{subsec:insights}).
For further validation, we create a dataset and settings where we have an approximated ground truth for \TCAVscore. We show that TCAV closely tracks the ground truth ~(Section~\ref{subsec:sanity}) while
saliency maps 
are unable to communicate this ground truth to humans  (Section~\ref{subsec:humanexp}). 
Finally we apply TCAV to help interpret a model predicting diabetic retinopathy (DR) (Section~\ref{subsec:retina}), where TCAV provided insights when the model diverged with the domain expert's knowledge.

\subsection{Validating the learned CAVs}

The first step is to convince ourselves that the learned CAVs are aligned with the intended concepts of interest. We first sort the images of any class with respect to CAVs for inspection. Then we learn patterns that maximally activate each CAV using an activation maximization technique for further visual confirmation. 
\subsubsection{Sorting images with CAVs\label{subsec:sorter}}

We can use CAVs to sort images with respect to their relation to the concept. This is useful for qualitative
confirmation that the CAVs correctly reflect the concept of interest.
As a CAV encodes the direction of a concept in the vector space of a bottleneck, $v_C^l \in \mathbb{R}^m$ using 
the activations of the concept pictures, $f_l(x_i) \in \mathbb{R}^m$ as described  Section~\ref{subsec:learnCAV}, 
we can compute cosine similarity between a set of pictures of interest to the CAV to sort the pictures. Note that the pictures being sorted are not used to train the CAV. 

\begin{figure}[h]
    \centering
    \includegraphics[width=1 \linewidth]{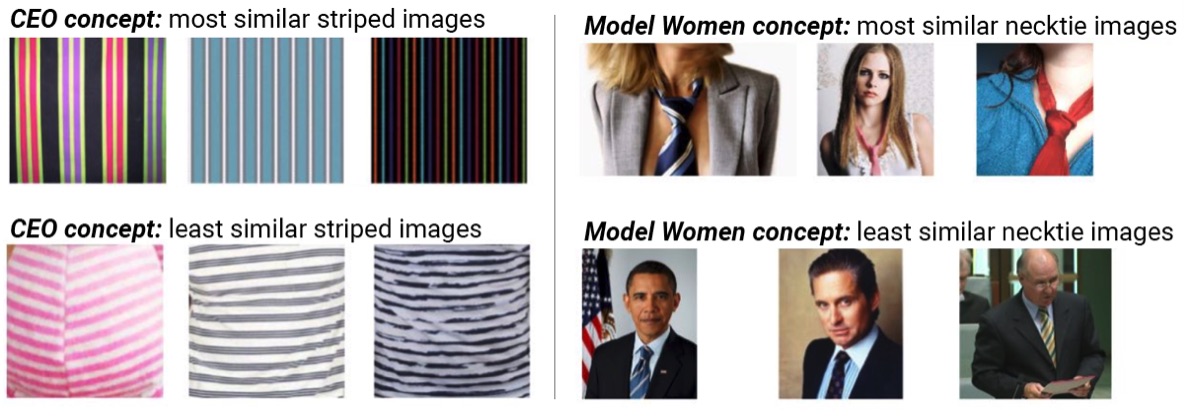}
    \vspace*{-5ex}
    \caption{The most and least similar pictures of stripes using `CEO' concept (left) and neckties using `model women' concept (right)}
    \label{fig:cosinesim}
\end{figure}
\vspace*{-2ex}

The left of Figure~\ref{fig:cosinesim} shows sorted images of stripes with respect to a CAV learned from a more abstract concept, `CEO' (collected from ImageNet).
The top 3 images are pinstripes which may relate to the ties or suits that a CEO may wear. 
The right of Figure~\ref{fig:cosinesim} shows sorted images of neckties with respect to a `model women' CAV. All top 3 images show women in neckties. 

This also suggests that CAVs can be as a standalone similarity sorter, to sort images to reveal any biases in the example images from which the CAV is learned.

\subsubsection{Empirical Deep Dream\label{subsec:deepdream}}

Another way to visually confirm our confidence in a CAV is to optimize for a pattern that maximally activates the CAV and compare that to our semantic notions of the concept. 
Activation maximization techniques, such as Deep Dream or Lucid~\citep{mordvintsev2015inceptionism, olah2017feature}, are
often used to visualize patterns that would maximally activate a neuron, set of neurons or random directions.
 This technique is also applied to AI-aided art~\citep{mordvintsev2015inceptionism}.
As is typically done, we use a random image as a starting point for the optimization to avoid choosing an arbitrary image.

\begin{figure}[h]
\centering
            \includegraphics[width=1. \linewidth]{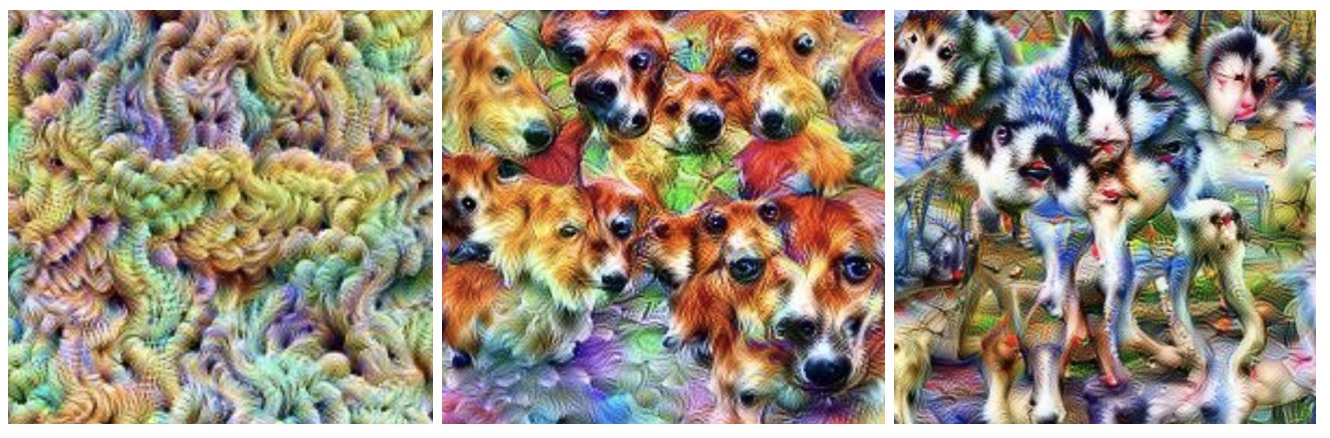}\vspace*{-4ex}
    \caption{Empirical Deepdream using knitted texture, corgis and Siberian huskey concept vectors (zoomed-in)}
    \label{fig:deepdream_indv}
\end{figure}

Using this technique, we show that CAVs do reflect their underlying concepts of interest.
Figure~\ref{fig:deepdream_indv} shows the results of deep dreamed patterns for knitted texture, corgis and Siberian huskey CAVs. We include
results from all layers and many other CAVs in the appendix.
This suggests that TCAV can be used to identify and visualize interesting directions in a layer. 
\vspace*{-2ex}
\vspace*{-3ex}

\subsection{Insights and biases: TCAV for widely used image classifications networks}

In this section, we apply TCAV to two popular networks to 1) further confirm TCAV's utility, 2) reveal biases, and 3) show where concepts are learned in these networks. 

\subsubsection{Gaining insights using TCAV\label{subsec:insights}}

We applied TCAV for two widely used networks~\cite{ szegedy2015going,szegedy2016rethinking}. We tried various types of CAVs, including color, texture, objects, gender and race. Note that none of these concepts were in the set of 
the network's class labels;
instead all were collected from~\cite{netdissect2017, huang2007labeled, russakovsky2015imagenet} or a popular image search engine. We show TCAV results with CAVs learned from all (for GoogleNet) or a subset (for Inception V3) of layers.

\begin{figure}[h]
    \centering
        \includegraphics[width=.88 \linewidth]{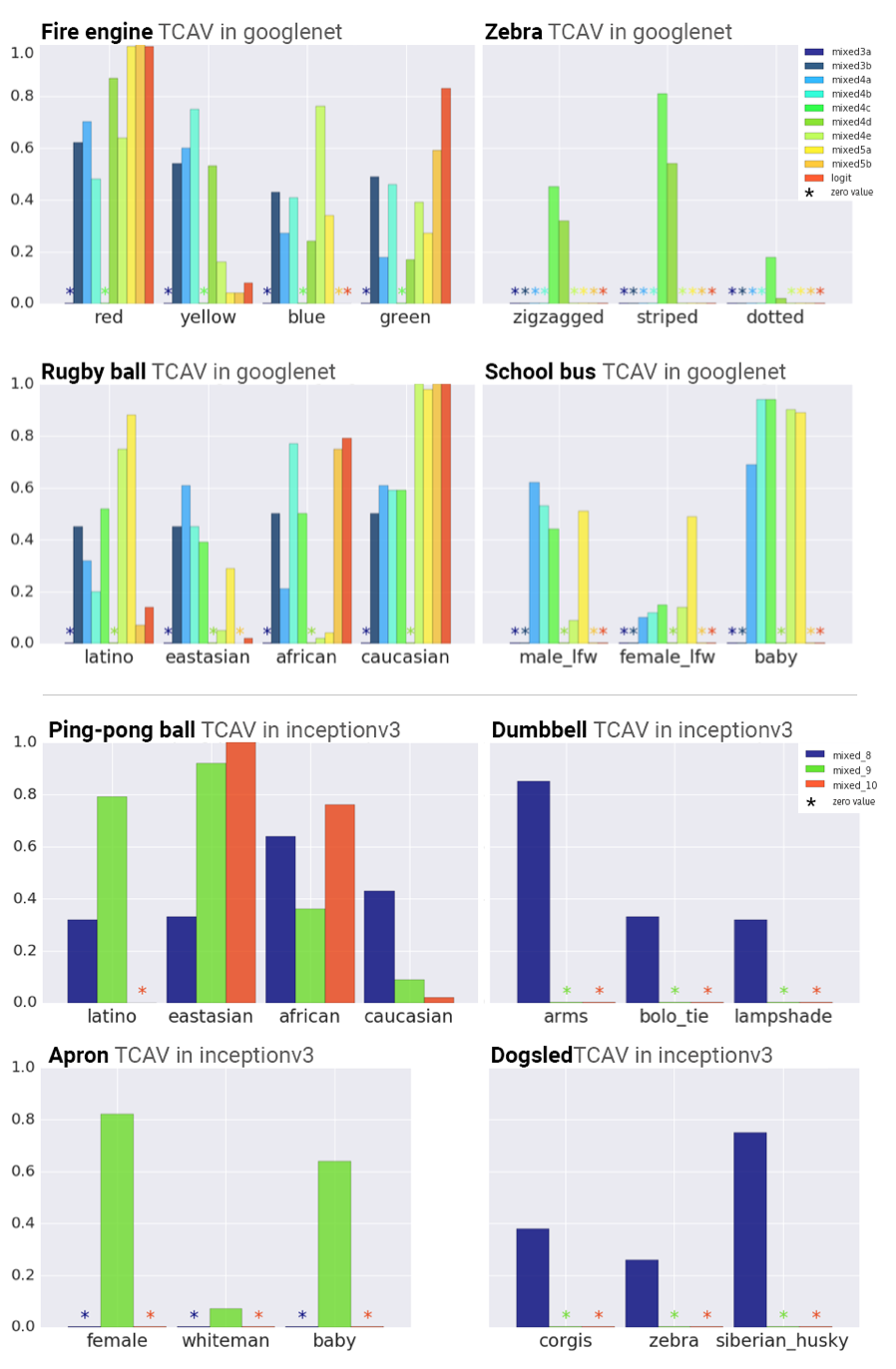}\vspace*{-1ex}
    \caption{Relative TCAV for all layers in GoogleNet~\cite{szegedy2015going} and
    last three layers in Inception V3 ~\cite{szegedy2016rethinking} for confirmation (e.g., fire engine), discovering biases (e.g., rugby, apron), and
   quantitative confirmation for previously qualitative findings in ~\cite{mordvintsev2015inceptionism, stock2017convnets} (e.g., dumbbell, ping-pong ball). \TCAVscore{}s~in layers close to the logit layer (red) represent more direct influence on the prediction than lower layers. `*'s mark CAVs omitted after statistical testing. }
    \label{fig:insights}
\end{figure}\vspace*{-1ex}

As shown in Figure~\ref{fig:insights}, some results confirmed our common-sense intuition, such as the importance of the red concept for fire engines, the striped concept for zebras, and the Siberian husky concept for dogsleds.
Some results also confirmed our suspicion that these networks were sensitive to gender and race, despite not being explicitly trained with these categories.
For instance, TCAV provides quantitative confirmations to the qualitative findings from~\cite{stock2017convnets} that found ping-pong balls are highly correlated with a particular race. TCAV also finds the `female' concept highly relevant to the `apron' class. Note that the race concept (ping-pong ball class) shows a stronger signal as it gets closer to the final prediction layer, while the texture concept (e.g., striped) influences ~\TCAVscore{}~in earlier layers (zebra class).

We also observed that the statistical significance testing (Section~\ref{sec:significance}) of CAVs successfully filters out spurious results. For instance, it successfully filtered out spurious CAVs where the `dotted' concept returned high \TCAVscore~(e.g., mixed4a) for zebra classes. The statistical significance testing of CAVs successfully eliminated CAVs in this layer; all CAVs that passed this testing consistently returned `striped' as the most important concept.

In some cases, it was sufficient to use a small number of pictures to learn CAVs. For the `dumbbell' class, we collected 30 pictures of each concept from a popular image search engine.
Despite the small number of examples, Figure~\ref{fig:insights} shows that TCAV successfully identified that the `arms' concept was more important to predict dumbbell class than other concepts. This finding is consistent with previous qualitative findings from~\cite{mordvintsev2015inceptionism}, where a neuron's DeepDream picture of a dumbbell showed an arm holding it. TCAV allows for quantitative confirmation of this previously qualitative finding.

\subsubsection{TCAV for where concepts are learned\label{subsec:wherelearned}}

In the process of learning CAVs, we train a linear classifier to separate each concept. We can use the performance of these linear classifiers to obtain lower-bound approximates for which layer
each concept is learned. 
\vspace*{-2ex}
\begin{figure}[h!]
    \centering
    \includegraphics[width=.9\linewidth]{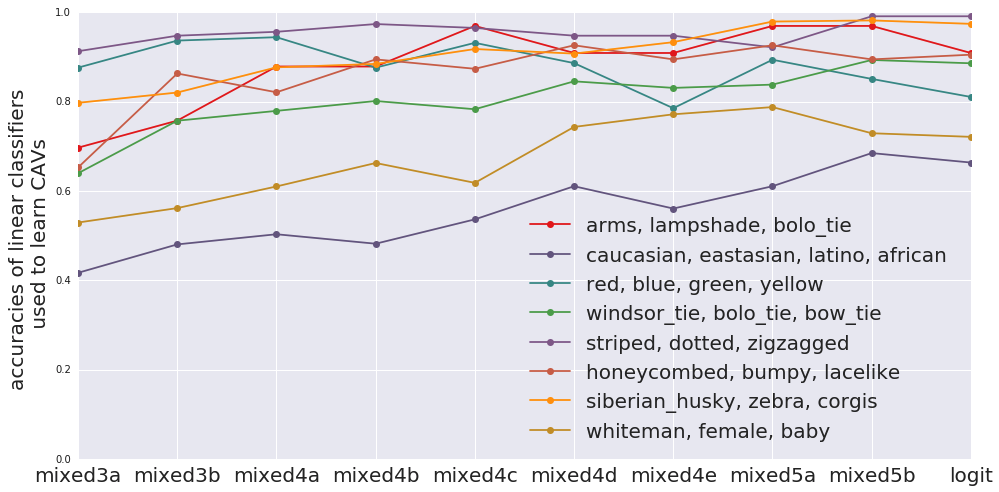}\vspace*{-2ex}
    \caption{The accuracies of CAVs at each layer. Simple concepts (e.g., colors) achieve higher performance
    in lower-layers than more abstract or complex concepts (e.g. people, objects)\label{fig:basis_acc} }
\end{figure}

Figure~\ref{fig:basis_acc} shows that the accuracy of more abstract concepts (e.g., objects) increases in higher layers of the network. The accuracy of simpler concepts, such as color, is high throughout the entire network. 
This is a confirmation of many prior findings~\cite{zeiler2014visualizing} that lower layers operate as lower level feature detectors (e.g., edges), while higher layers use these combinations of lower-level features to infer higher-level features (e.g., classes).
The accuracies are measured by a held out test set of 1/3 the size of the training set. 

\subsection{A controlled experiment with ground truth\label{subsec:sanity}}

\vspace*{-2ex}
\begin{figure} [h!]
\centering
\includegraphics[width=.5\textwidth]{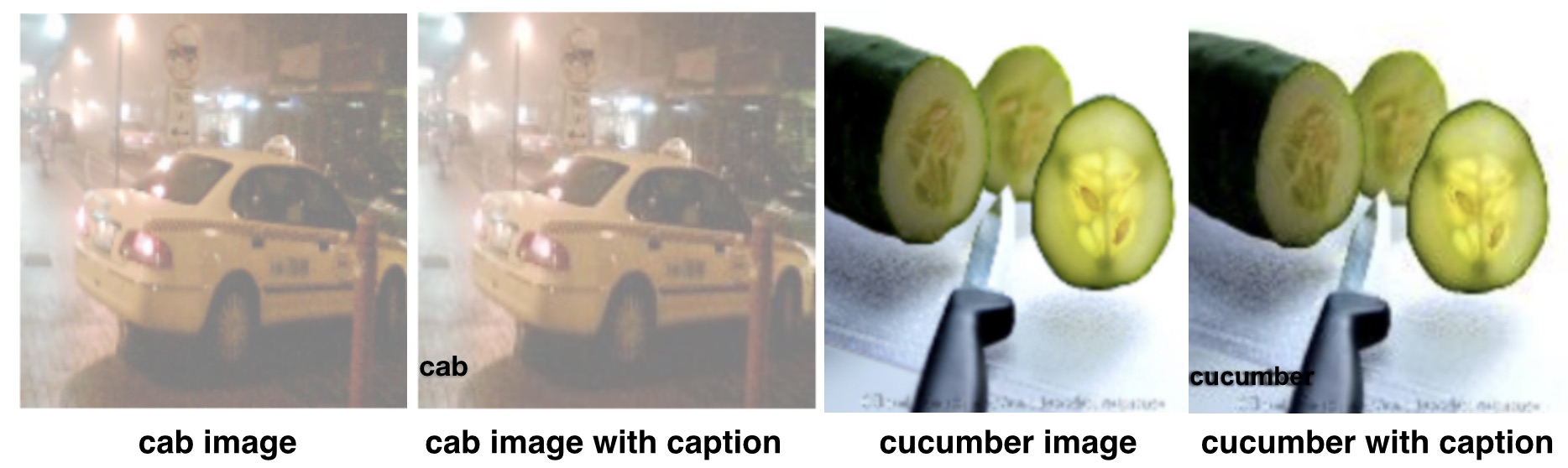}
\vspace*{-4ex}
\caption{A controlled training set: Regular images and images with captions for the cab and cucumber class.\label{fig:setup}}
\end{figure}

The goal of this experiment is demonstrate that TCAV can be successfully used to interpret the function learned by a neural network in a carefully controlled setting where ground truth is known. We show quantitative results of TCAV and compare these with our evaluation of saliency maps.

To this end we create a dataset of three arbitrary classes (zebra, cab, and cucumber)
with potentially noisy captions written in the image (example shown in Figure~\ref{fig:setup}). The noise parameter $p \in [0, 1.0]$ controls the probability that the image caption agrees with the image class. If there is no noise ($p = 0$), the caption always agrees with the image label, e.g. a picture of a cab always contains the word ``cab'' at the bottom. At $p=.3$, each picture has a 30\% chance of having the correct caption replaced with a random word (e.g. ``rabbit'').

We then train 4 networks, each on a dataset with a different noise parameter $p$ in $[0,1]$. Each network may learn to pay attention to either images or captions (or both) in the classification task. To obtain an approximated ground truth for which concept each network paid attention, we can test the network's
performance on images without captions. If the network used the image concept for classification, the performance should remain high. If not, the network performance will suffer. We create image CAVs using each class's images, and caption CAVs using captions with other pixels in the image randomly shuffled.

\subsubsection{Quantitative evaluation of TCAV \label{subsec:sanity}}

\begin{figure}[h!]
\centering
\includegraphics[width=.49 \linewidth]{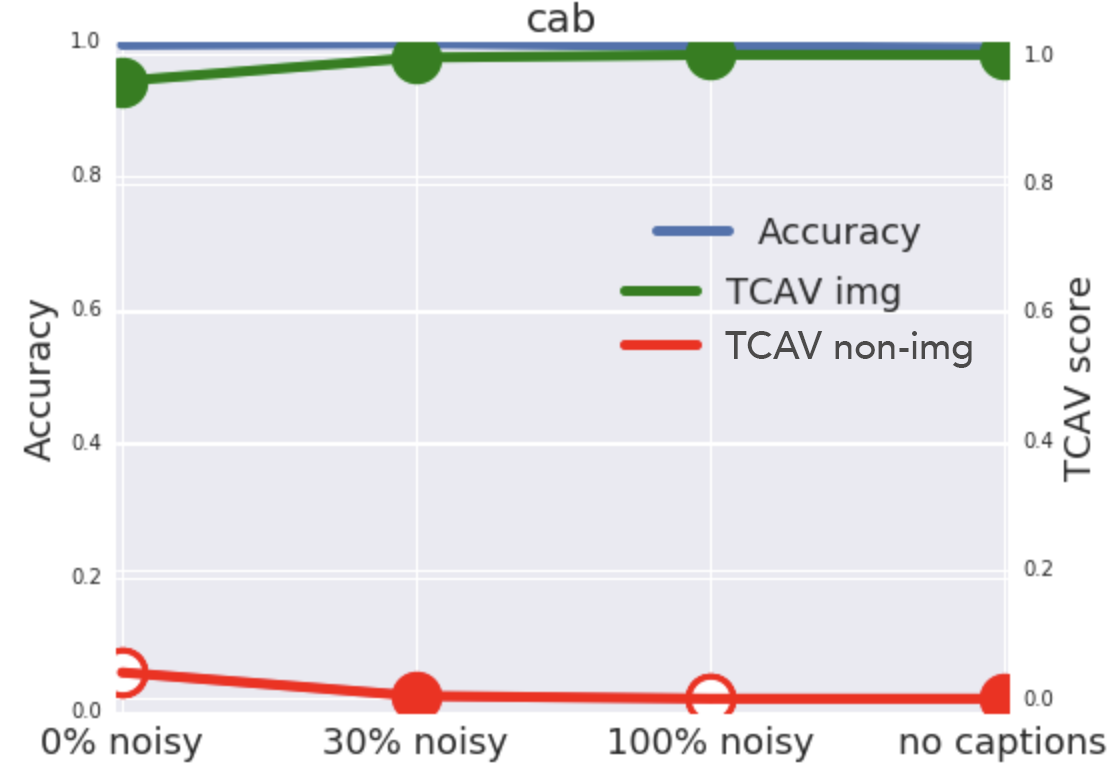}
\includegraphics[width=.49\linewidth]{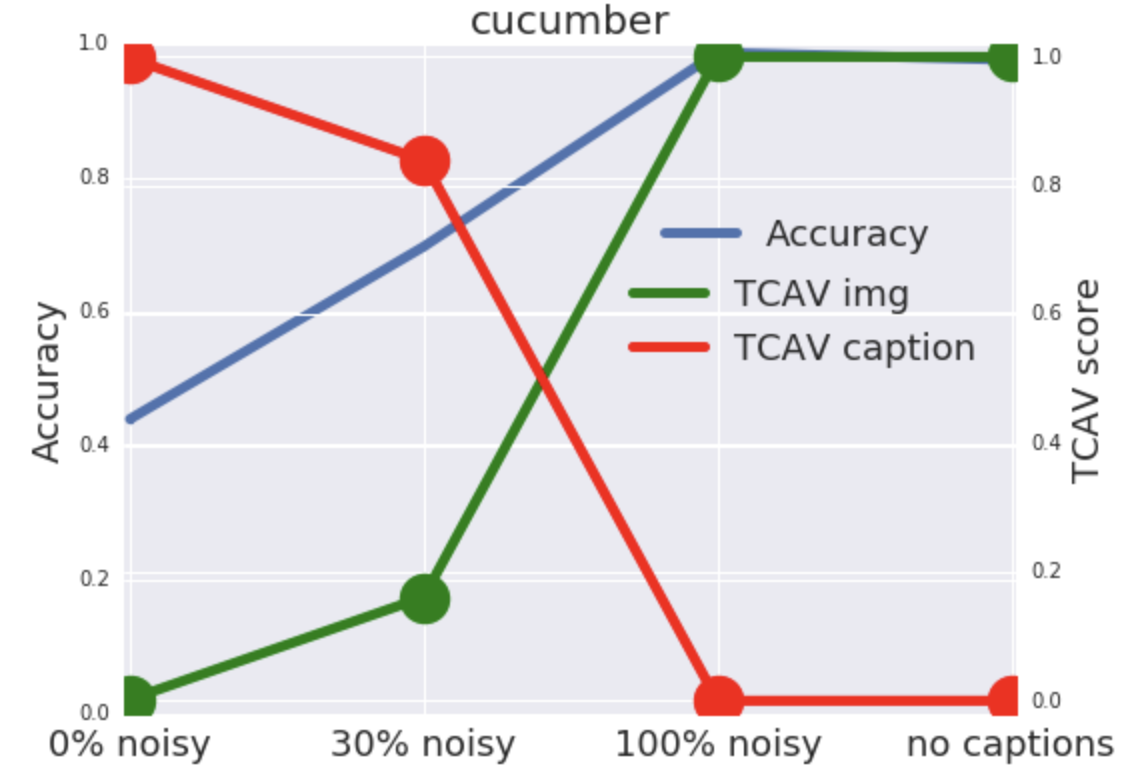} 
\caption{TCAV results with approximated ground truth: 
Both cab and cucumber classes, \TCAVscore{}~closely matches the ground truth. For the cab class, the network used image concept more than the caption concept regardless of the models. 
\label{fig:sanity_tcav}}
\end{figure}\vspace*{-2.5ex}

Overall, we find that the TCAV score closely mirrors the concept that the network paid 
attention to (Figure~\ref{fig:sanity_tcav}). Accuracy results suggest that, when classifying cabs, the network used the image concept more than the caption concept, regardless of the noise parameter.
However, when classifying cucumbers, the network sometimes paid attention to the caption concept and sometimes the image concept. 
Figure~\ref{fig:sanity_tcav} shows that the \TCAVscore~closely matches this ground truth. In the cab class, the \TCAVscore~for the image concept is high, consistent with its high test performance on caption-less images. In the cucumber class, the \TCAVscore~for the image concept increases as noise level increases, consistent with the observation that accuracy also increases as noise increases.

\subsubsection{Evaluation of saliency maps with human subjects\label{subsec:humanexp}}

\begin{figure}
\centering
\includegraphics[width=.9\linewidth]{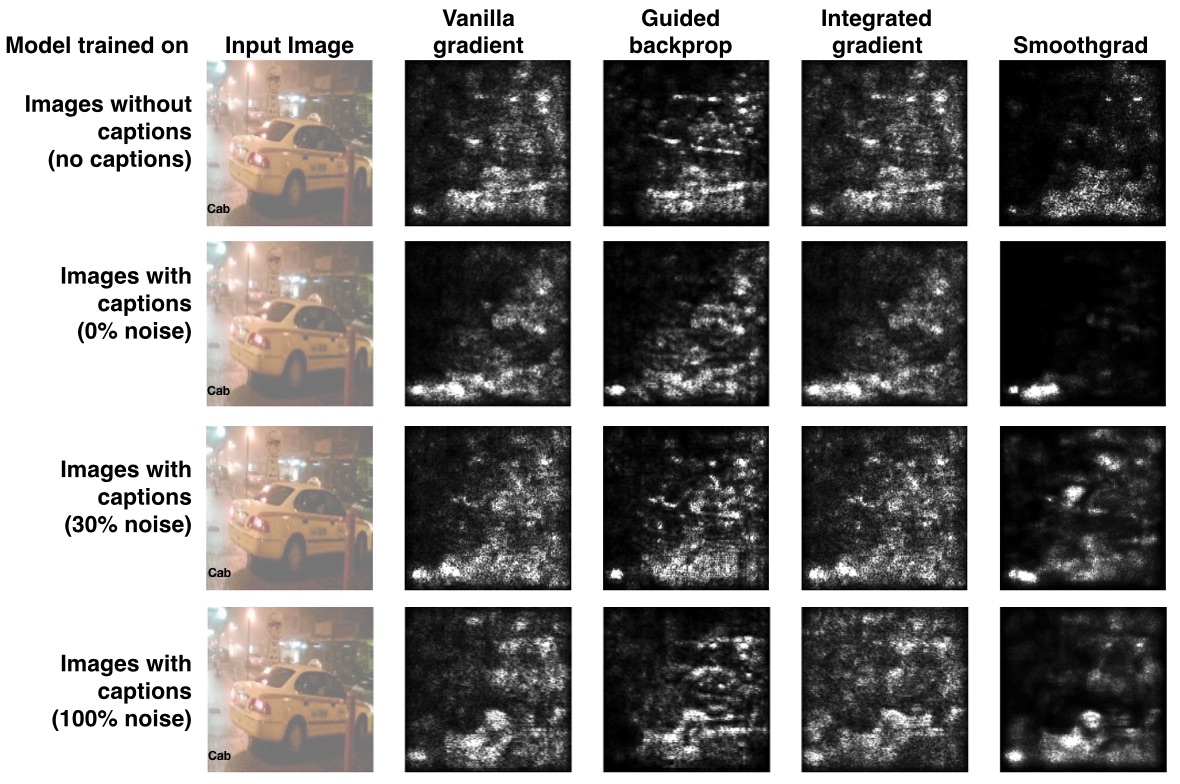}
\vspace*{-2ex}
\caption{Saliency map results with approximated ground truth: Models trained on datasets with different noise parameter $p$ (rows) and different saliency map methods (columns) are presented. The approximated ground truth is that the network is paying a lot more attention to the image than the caption in all cases, which is not clear from saliency maps.  \label{fig:sanity_sal}}
\end{figure}

Saliency maps are an alternative way to communicate the same information, and are commonly used as an interpretability method for image-based networks (see Section~\ref{sec:related}). Qualitatively, as shown in Figure~\ref{fig:sanity_sal} for the cab class, it is not clear that the four networks used the image concept more than the caption concept. 
In this section, we quantitatively evaluate what information saliency maps are able to communicate to humans, via a human subject experiment.

We took the saliency maps generated from the previous section to conduct a 50-person human experiment on Amazon Mechanical Turk. For simplicity, we evaluated two of the four noise levels (0\% and 100\% noise), and two types of saliency maps (\cite{sundararajan2017axiomatic} and~\cite{smilkov2017smoothgrad}).

Each worker did a series of six tasks (3 object classes $\times$ 2 saliency map types), all for a single model. Task order was randomized. In each task, the worker first saw four images along with their corresponding saliency masks. They then rated how important they thought the image was to the model (10-point scale), how important the caption was to the model (10-point scale), and how confident they were in their answers (5-point scale). In total, turkers rated 60 unique images (120 unique saliency maps). 

\begin{figure}
\centering
\includegraphics[width=.9\linewidth]{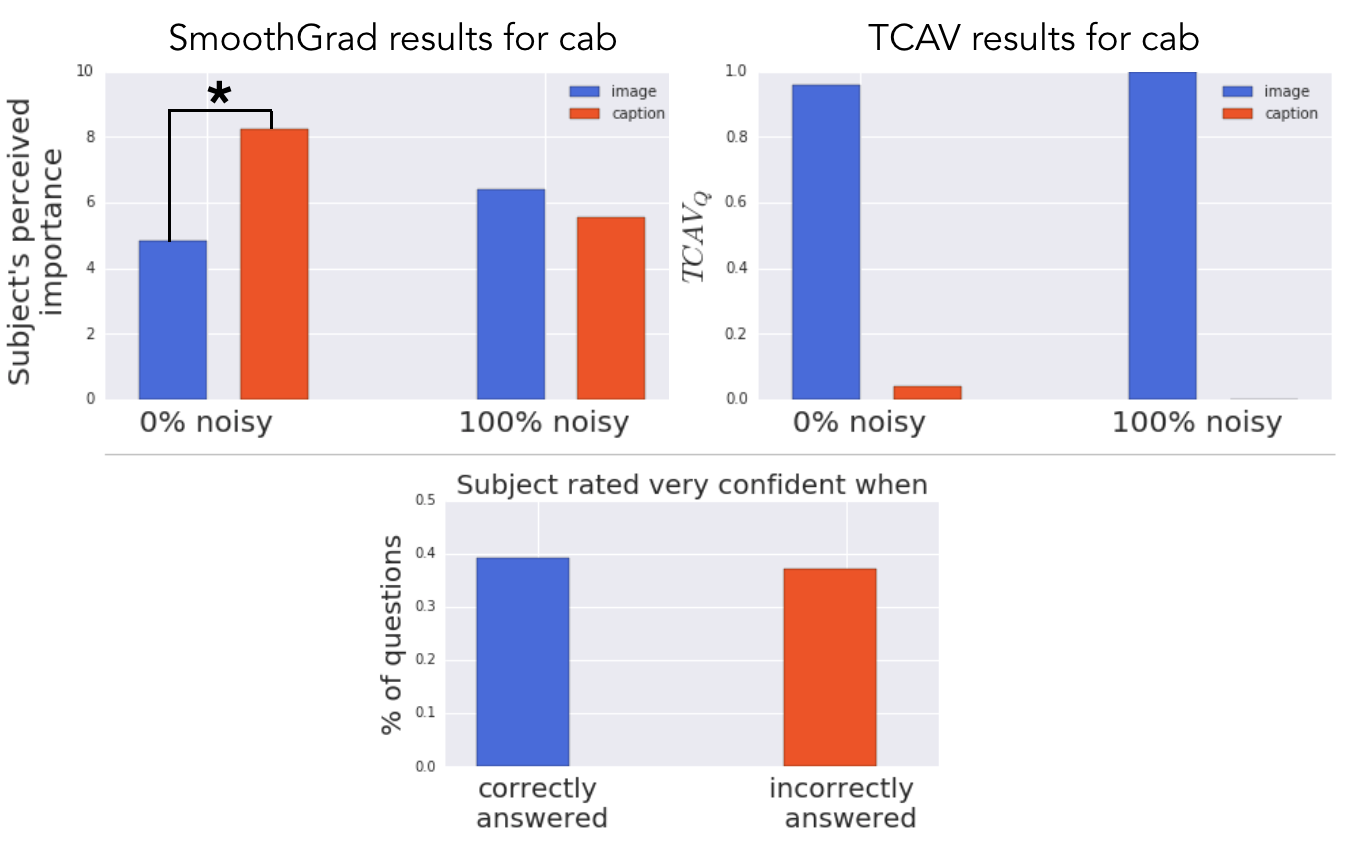}
\vspace*{-2ex}
\caption{For the cab class, the ground truth was that the image concept was more important than the caption concept. However, when looking at saliency maps, humans perceived the caption concept as being more important (model with 0\% noise), or did not discern a difference (model with 100\% noise). In contrast, TCAV results correctly show that the image concept was more important. Overall, the percent of correct answers rated as very confident was similar to that of incorrect answers, indicating that saliency maps may be misleading.\label{fig:human_exp_simple}}\vspace*{-2ex}
\end{figure} 

Overall, saliency maps correctly communicated which concept was more important only 52\% of the time (random chance is 50\% for two options). Wilcox signed-rank tests show that in more than half of the conditions, there was either no significant difference in the perceived importance of the two concepts, or the wrong concept was identified as being more important. Figure~\ref{fig:human_exp_simple} (top) shows one example where saliency maps communicated the wrong concept importance. In spite of this, the percent of correct answers rated as very confident was similar to that of incorrect answers (Figure~\ref{fig:human_exp_simple} bottom), suggesting that interpreting using saliency maps alone could be misleading. Furthermore, when one of the saliency map methods correctly communicated the more important concept, it was always the case that the other saliency map method did not, and vice versa.

\subsection{TCAV for a medical application\label{subsec:retina}}

\begin{figure}
\centering
\includegraphics[width=.8 \linewidth]{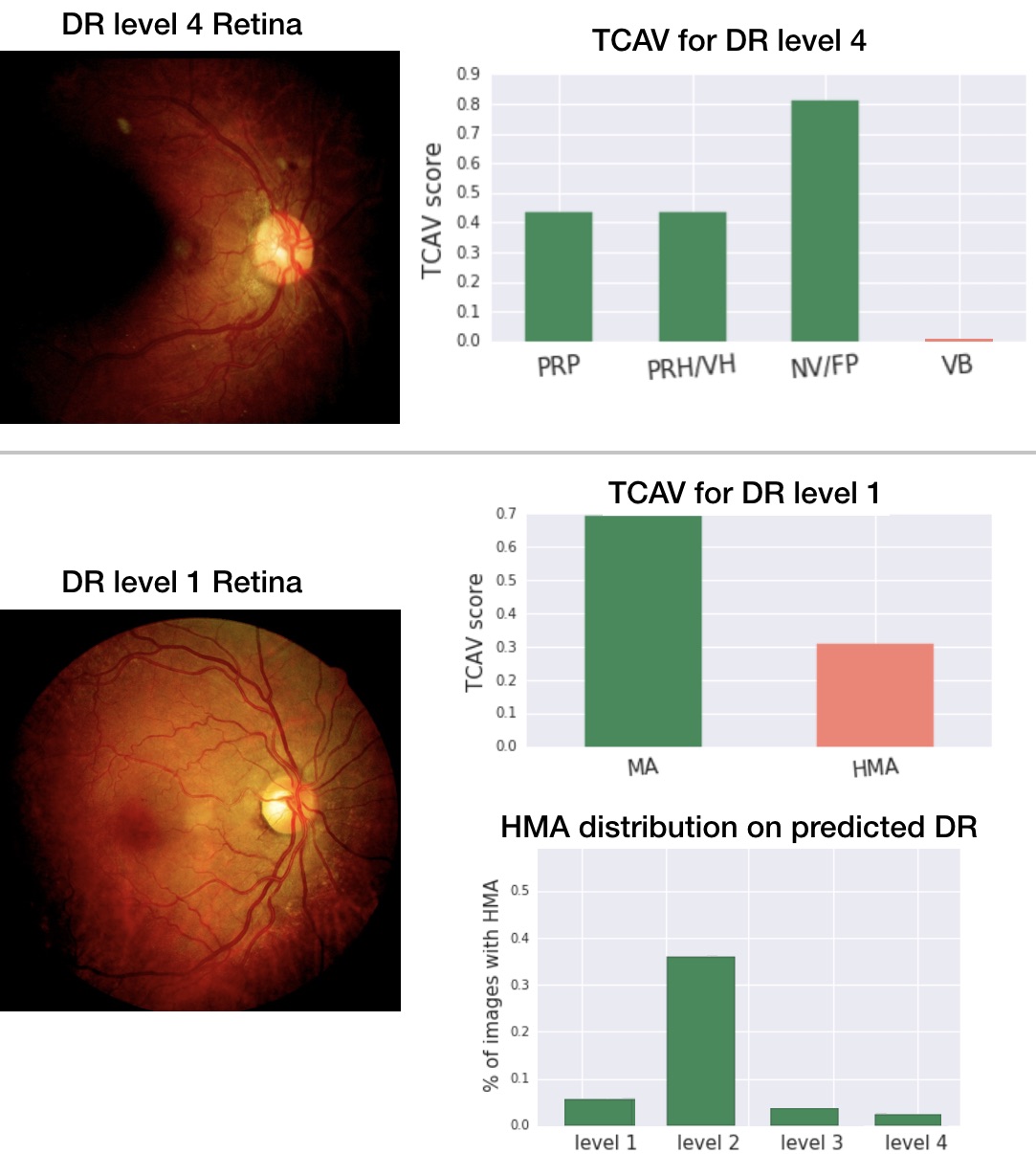}\vspace*{-2.3ex}
\caption{Top: A DR level 4 image and TCAV results. \TCAVscore~is high for features relevant for this level (green), and low for an irrelevant concept (red). 
Middle: DR level 1 (mild) TCAV results. The model often incorrectly predicts level 1 as level 2, a model error that could be made more interpretable using TCAV: \TCAVscore s on 
concepts typically related to level 1 (green, MA) are high in addition to level 2-related concepts (red, HMA). Bottom: the HMA feature appears more frequently in DR level 2 than DR level 1.
\label{fig:retina}}
\end{figure}\vspace*{-1ex}

We now apply TCAV to the real-world problem of predicting diabetic retinopathy (DR), a treatable but sight-threatening condition, from retinal fundus images~\cite{retina_data_paper}. We consulted with a medical expert about our results.

The model of interest 
predicts DR level 
using a 5-point grading scale based on complex criteria, from level 0 (no DR) to 4 (proliferative).
Doctors' diagnoses of DR level depend on evaluating a set of diagnostic concepts, such as microaneurysms (MA) or pan-retinal laser scars (PRP), with different concepts being more prominent at different DR levels. We sought to test the importance of these concepts to the model using TCAV. 

For some DR levels, TCAV identified the correct diagnostic concepts as being important. As shown in Figure~\ref{fig:retina} (top), the TCAV score was high for concepts relevant to DR level 4, and low for a non-diagnostic concept.

For DR level 1, TCAV results sometimes diverge from doctors' heuristics ( Figure~\ref{fig:retina} bottom). For example, aneurysms (HMA) had a relatively high TCAV score, even though they are diagnostic of a higher DR level (see HMA distribution in Figure~\ref{fig:retina}). However, consistent with this finding, the model often over-predicted level 1 (mild) as level 2 (moderate). Given this, the doctor said she would like to tell the model to de-emphasize the importance of HMA for level 1. Hence, TCAV may be 
useful for helping experts interpret and fix model errors when they disagree with model predictions.  

\section{Conclusion and Future Work}

The method presented here, TCAV, is a step toward creating a human-friendly linear interpretation of the internal state of a deep learning model, so that questions about model decisions may be answered in terms of natural high-level concepts. Crucially, these concepts do not need to be known at training time, and may easily be specified during a post hoc analysis via a set of examples.

Our experiments suggest TCAV can be a useful technique in an analyst's toolbox. We provided evidence that CAVs do indeed correspond to their intended concepts. We then showed how they may be used to give insight into the predictions made by various classification models, from standard image classification networks to a specialized medical application.

There are several promising avenues for future work based on the concept attribution approach. While we have focused on image classification systems, applying TCAV to other types of data (audio, video, sequences, etc.) may yield new insights. TCAV may also have applications other than interpretability:
for example, in identifying adversarial examples for neural nets (see appendix).
Finally, one could ask for ways to identify concepts automatically
and for a network that shows super-human performance, concept attribution may help humans improve their own abilities.

\label{submission}

\subsubsection*{Acknowledgments}
We would like to thank Daniel Smilkov for helpful discussions. We thank Alexander Mordvintsev for providing tfzoo code. 
We also thank Ethan R Elenberg, David Alvarez Melis and  an anonymous reviewer for helpful comments and discussions. 
We thank Alexander Mordvintsev, Chris Olah and Ludwig Schubert for generously allowing us to use their code for DeepDream.
Thanks to Christopher for sharing early work on doing attribution to semantically meaningful channels.
Work from Nicholas Carlini, on training linear classifiers for non-label concepts on logit-layer activations, was one of our motivations.
Finally, we would like to thank Dr. Zahra Rastegar for evaluating diabetic retinopathy results, and provided relevant medical expertise.

\clearpage

\appendix
\section*{Appendix}

In this appendix, we show other experiments we conducted and additional results and figures. 
\section{TCAV on adversarial examples\label{subsec:adv}}

\begin{figure}[h!]
    prediction: zebra\hspace{ .4in}  zebra \hspace{ .6 in} zebra \\
        \centering
            \includegraphics[width=.24 \linewidth]{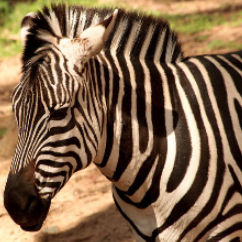}
        \includegraphics[width=.24 \linewidth]{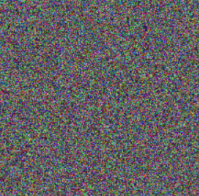}
        \includegraphics[width=.24 \linewidth]{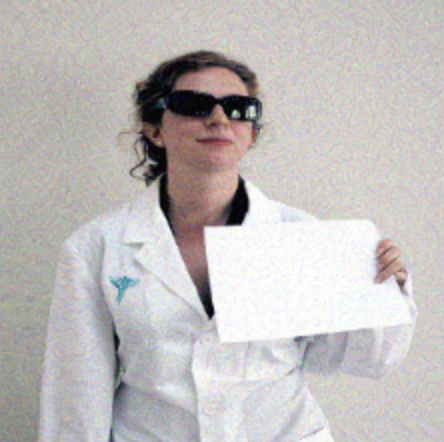}
        \includegraphics[width=.9 \linewidth]{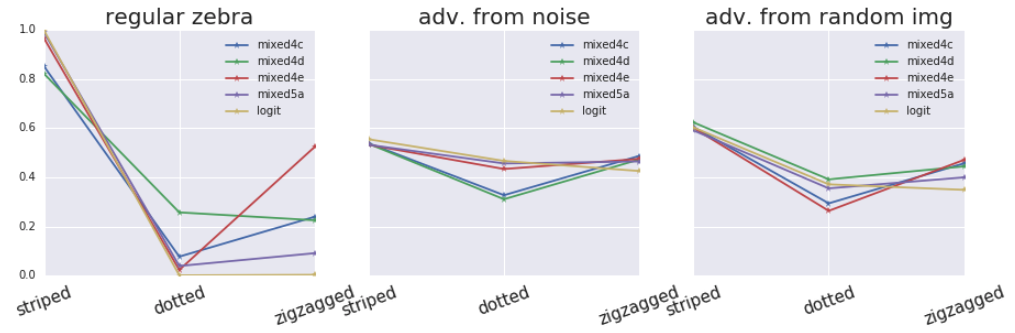}\vspace*{-2ex}
    \caption{Two types of adversarial images that are classified as zebra. In both cases, the distribution of \TCAVscore are different from that of a normal zebra.}
    \label{fig:adv1}
\end{figure}

Adversarial examples~\cite{szegedy2013intriguing} are small, often visually imperceptible changes to an image which can cause an arbitrarily change to a network's predicted class. We conduct a simple experiment to see whether TCAV is fooled by adversarial examples. In Figure~\ref{fig:adv1}, TCAV returns a high score for the striped concept for zebra pictures. We create two sets of adversarial examples, both by performing a targeted attack using a single step of the Fast Gradient Sign Method~\cite{alex17}. 
We successfully make the network believe that an essentially random noise image (top middle in Figure~\ref{fig:adv1}) 
 and a randomly sampled non-zebra image with imperceptible changes (top right in Figure~\ref{fig:adv1}) are zebras (100\% for noise image, 99\% for the latter).
However, the distribution of \TCAVscore s for the regular zebra and adversarial images remain different (bottom in Figure~\ref{fig:adv1}). 
While this may not be sufficient direction for building a defense mechanism, 
one can imagine having a dictionary of concepts where we know the usual distribution of \TCAVscore s for each class. If we want an 'alert' for potential adversarial attacks, 
we can compare the 'usual' distribution of \TCAVscore s to that of the suspicious images.

\section{Additional Results: Insights and biases: TCAV for widely used image classifications networks
}

We provide further results on Section 4.2.1. 
\begin{figure*}[h!]
        \centering
            \includegraphics[width=1\linewidth]{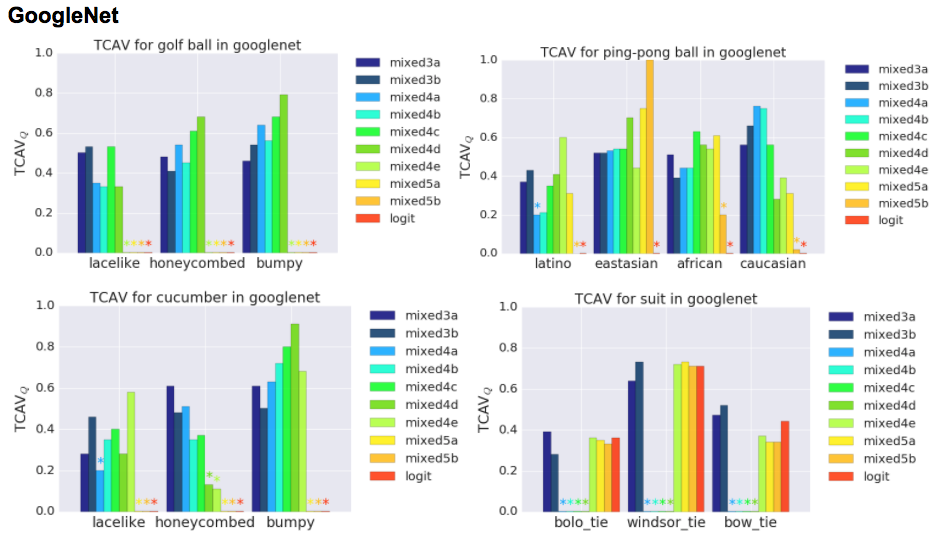}
       \includegraphics[width=.9 \linewidth]{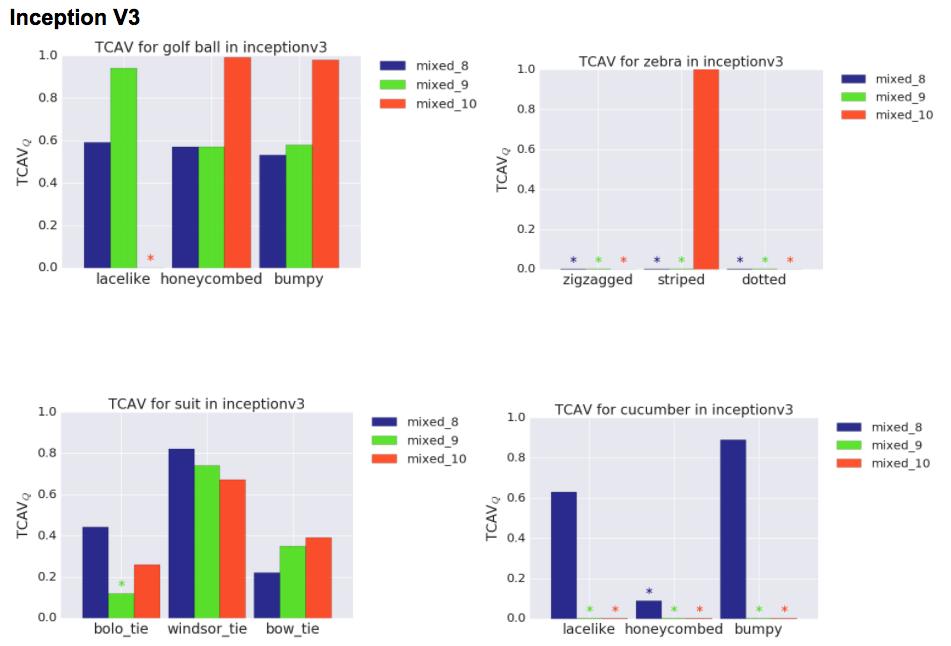}
    \caption{TCAV results for each layer}
    \label{fig:adv2}
\end{figure*}

\section{Additional Results: Empirical Deep Dream}

\begin{figure*}[h!]
        \centering
        \includegraphics[width=.9 \linewidth]{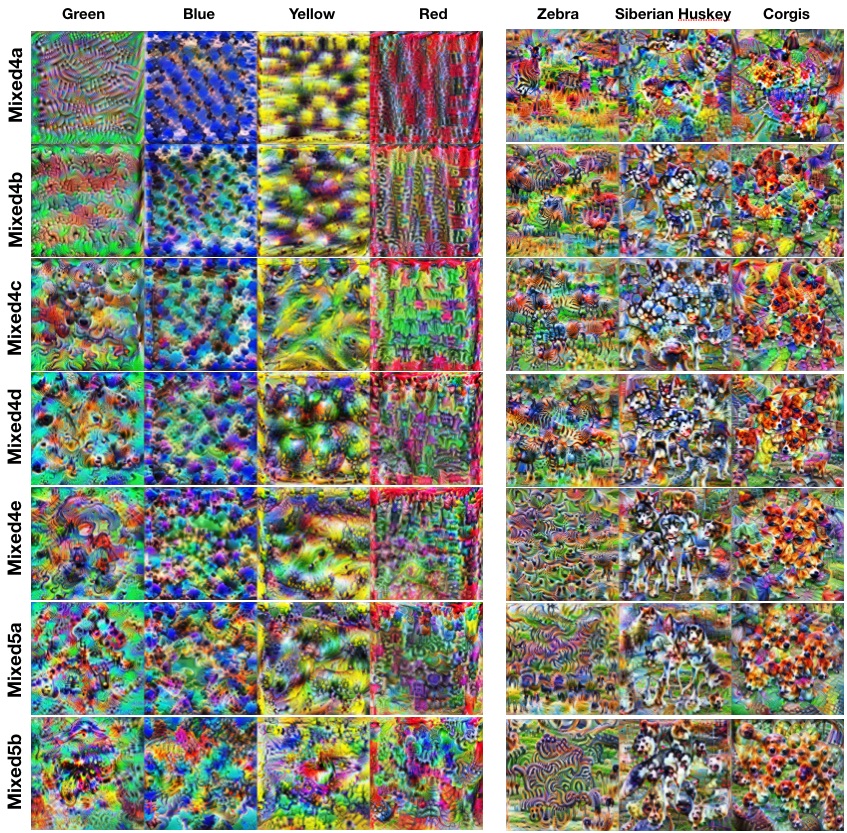}
    \caption{Empirical deepdream using CAVs for each layer in Googlenet.}
    \label{fig:adv3}
\end{figure*}

\begin{figure*}[h!]
        \centering
        \includegraphics[width=.9 \linewidth]{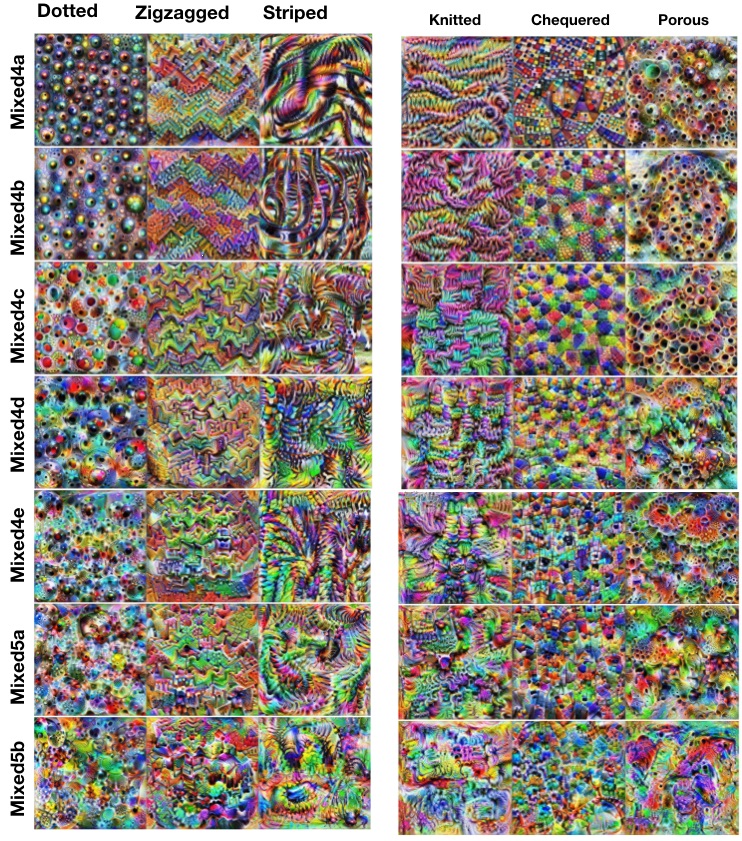}
    \caption{Empirical deepdream using CAVs for each layer in Googlenet.}
    \label{fig:adv4}
\end{figure*}

\section{Additional Results: Sorting Images with CAVs}

\begin{figure}[h!]
\centering
\includegraphics[width=.7\linewidth]{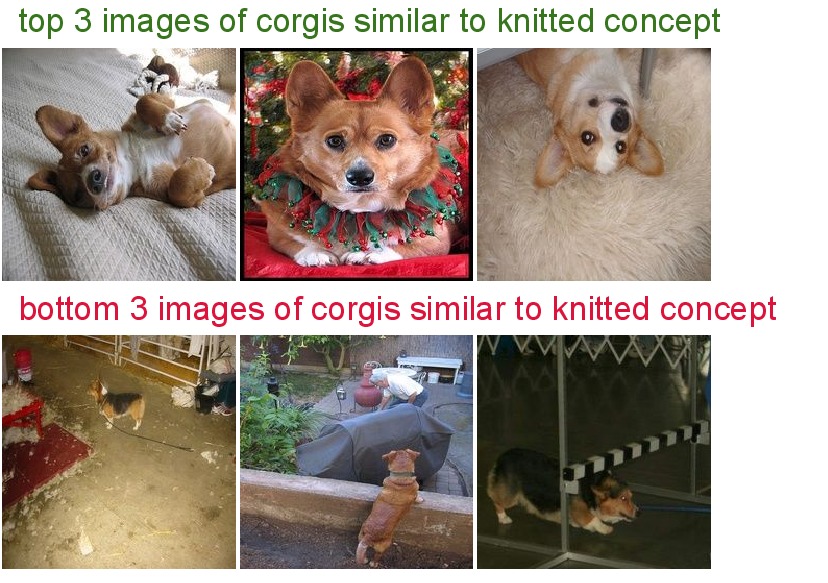}
\includegraphics[width=.7\linewidth]{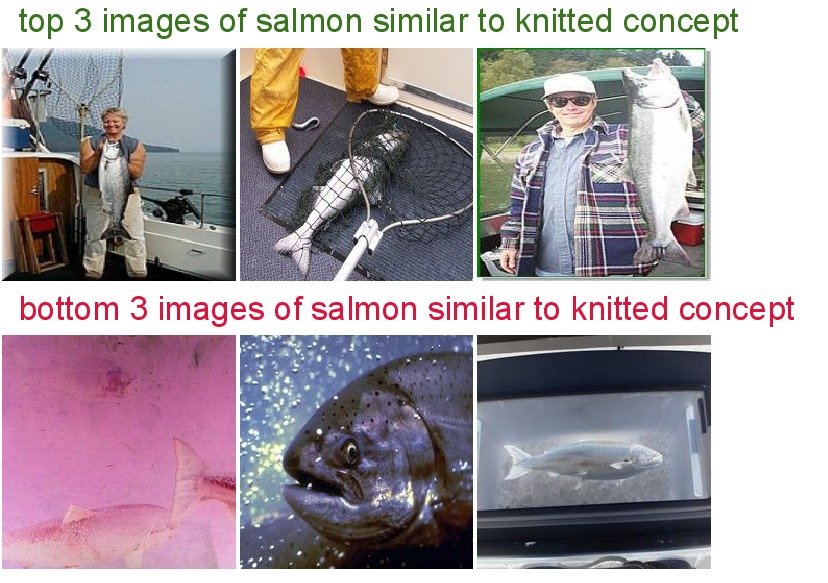}
\includegraphics[width=.7\linewidth]{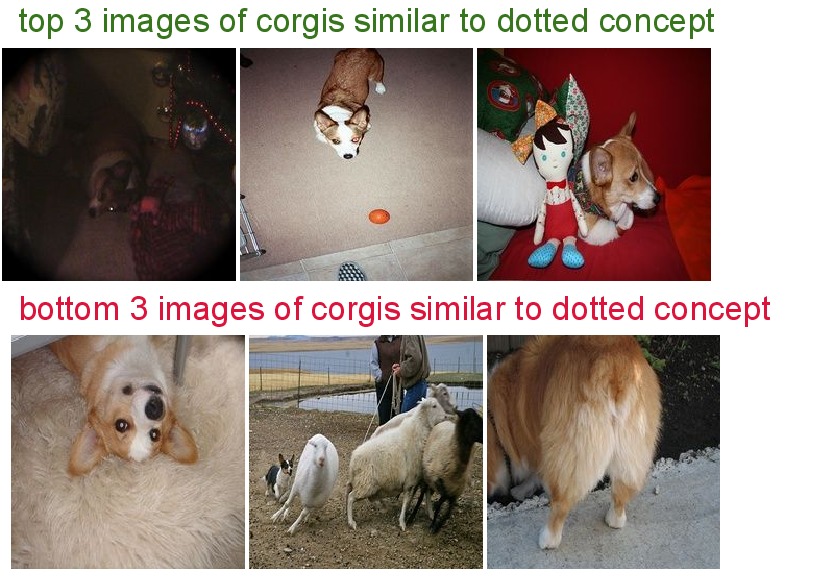}
\caption{Additional Results: Sorting Images with CAVs}
\end{figure}
\begin{figure}[h!]
\centering
\includegraphics[width=.7\linewidth]{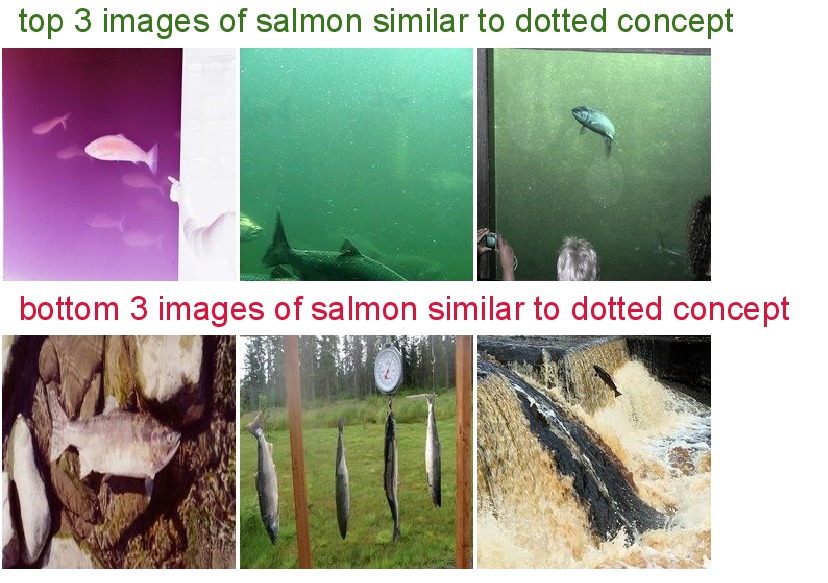}
\includegraphics[width=.7\linewidth]{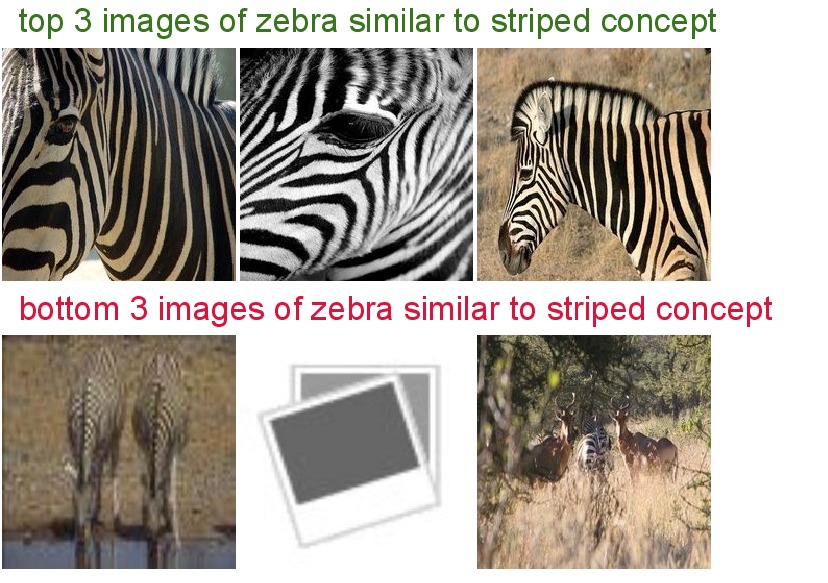}
\includegraphics[width=.7 \linewidth]{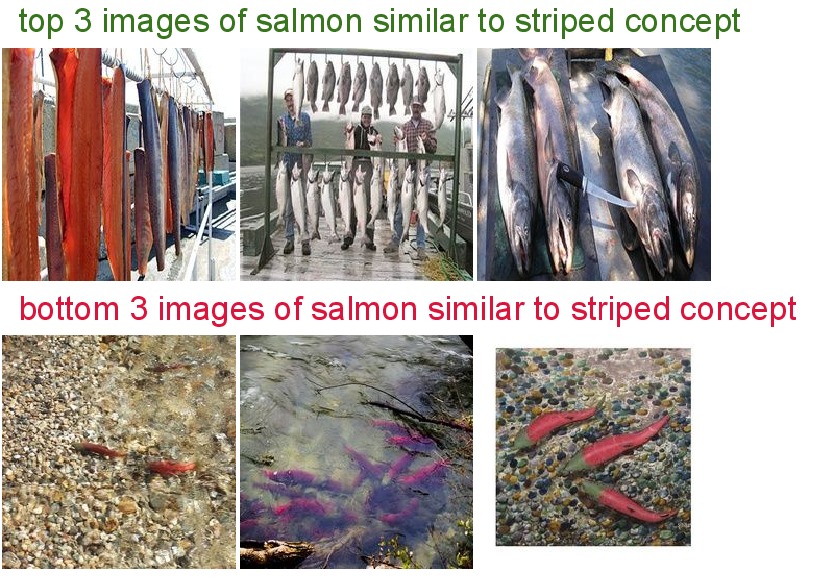}
\caption{Additional Results: Sorting Images with CAVs}

\end{figure}
\begin{figure}[h!]
\centering
\includegraphics[width=1 \linewidth]{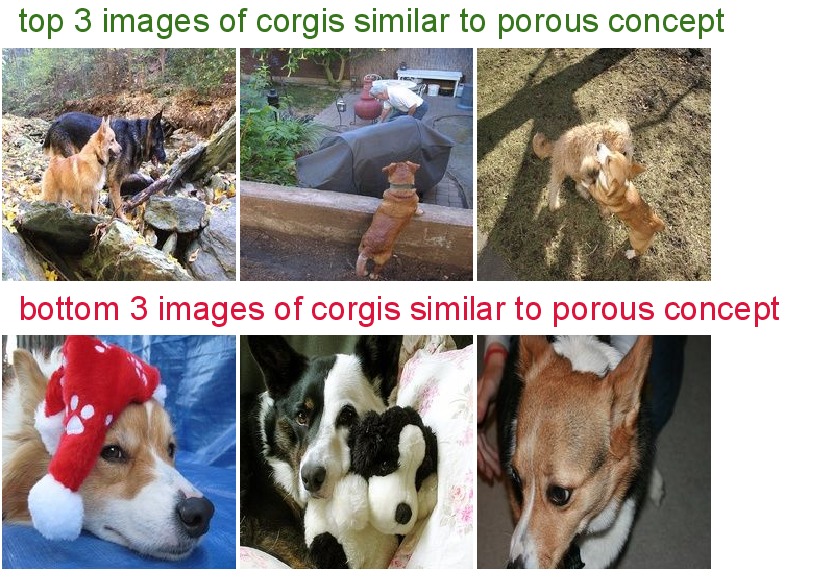}
\includegraphics[width=1 \linewidth]{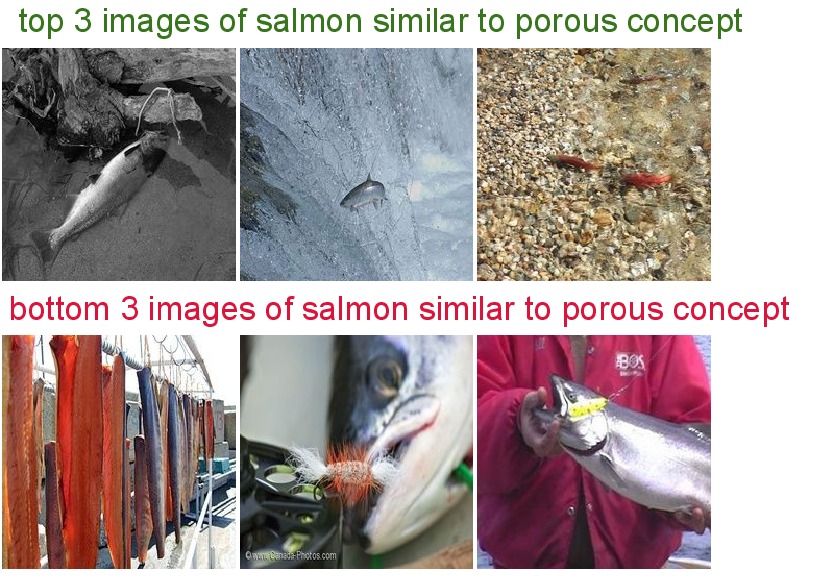}
\includegraphics[width=1 \linewidth]{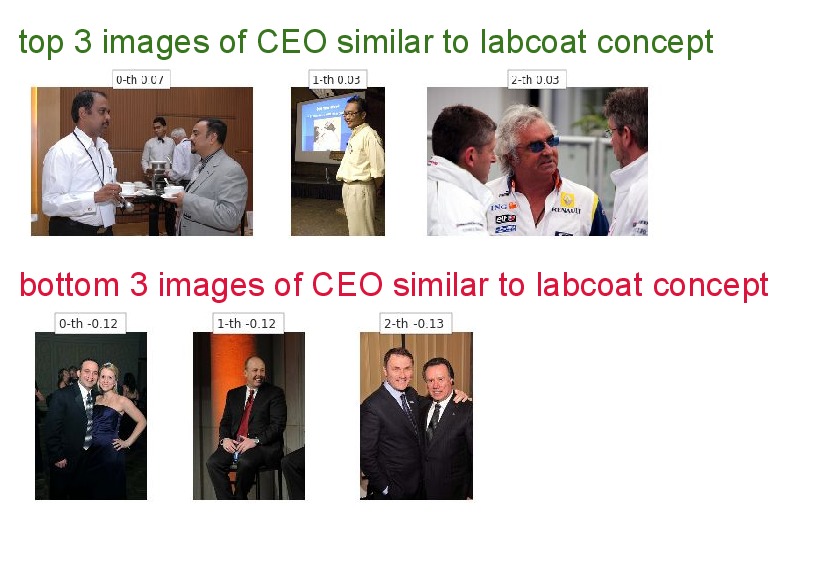}
\caption{Additional Results: Sorting Images with CAVs}

\end{figure}
\begin{figure}[h!]
\centering
\includegraphics[width=1 \linewidth]{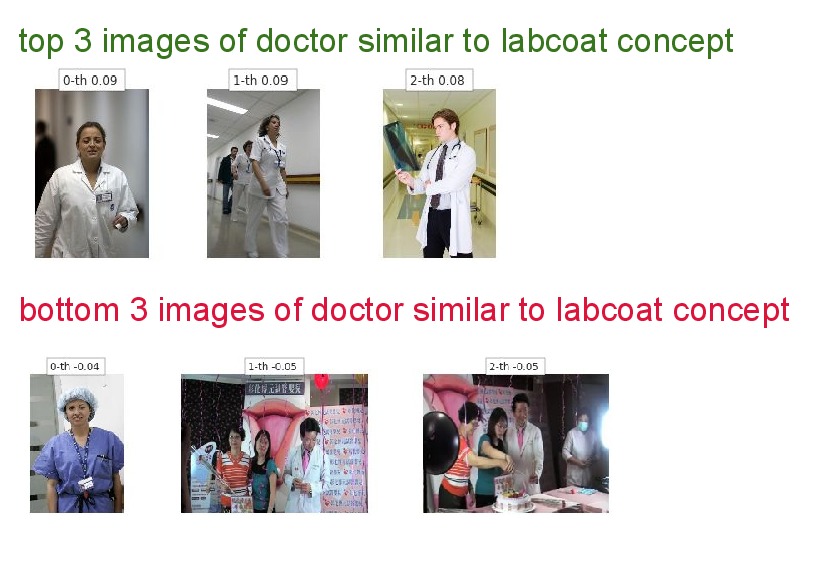}
\includegraphics[width=1 \linewidth]{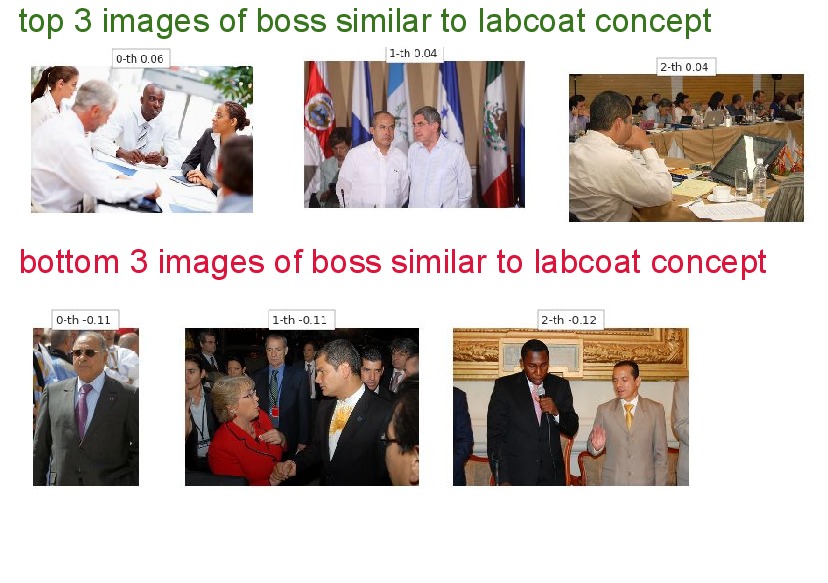}
\includegraphics[width=1 \linewidth]{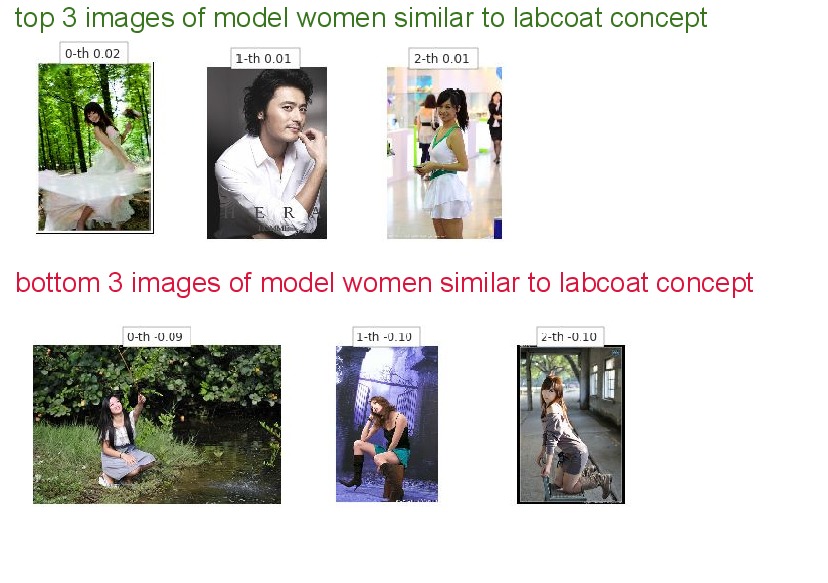}
\caption{Additional Results: Sorting Images with CAVs}

\end{figure}
\begin{figure}[h!]
\centering
\includegraphics[width=1 \linewidth]{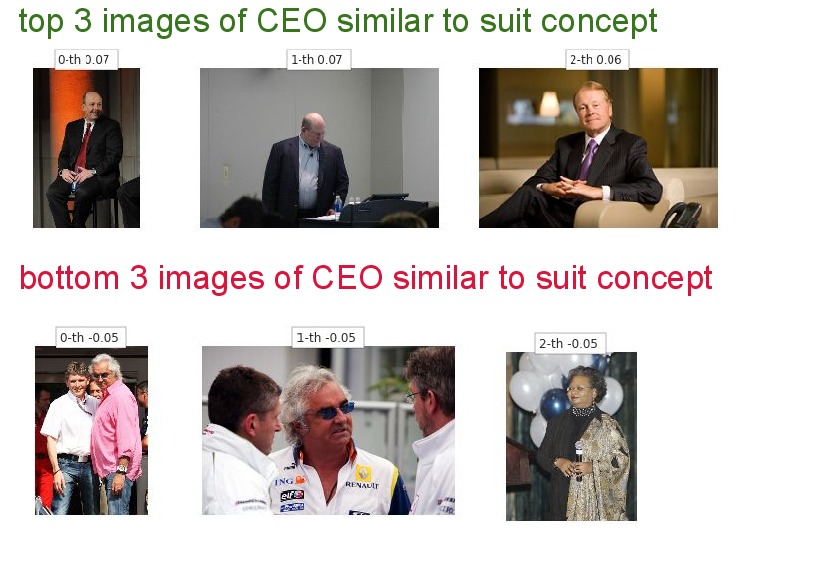}
\includegraphics[width=1 \linewidth]{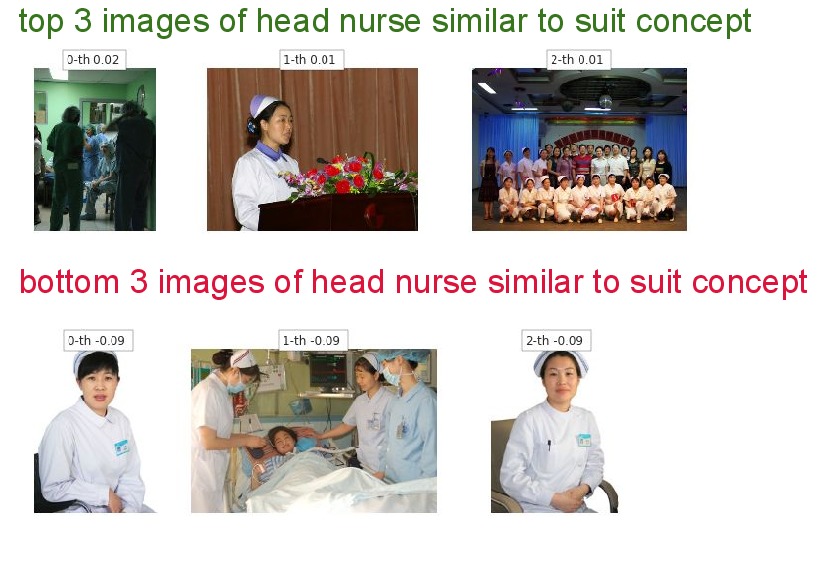}
\includegraphics[width=1 \linewidth]{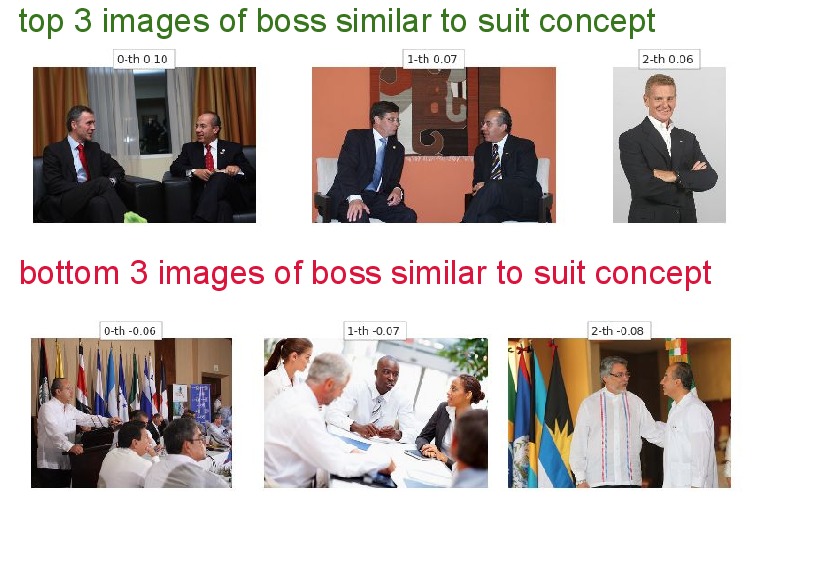}
\includegraphics[width=1 \linewidth]{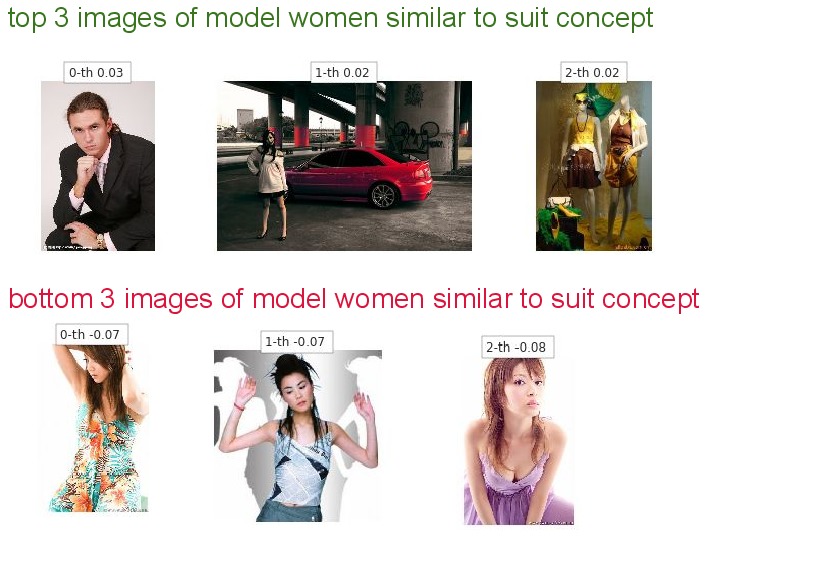}
\caption{Additional Results: Sorting Images with CAVs}

\end{figure}

\clearpage

\bibliography{main_from_Been}
\bibliographystyle{icml2018}

\end{document}